\documentclass[journal]{IEEEtran}
\usepackage{amsmath,amsfonts}
\usepackage{array}
\usepackage{textcomp}
\usepackage{stfloats}
\usepackage{url}
\usepackage{verbatim}
\usepackage{graphicx,color,xcolor}
\usepackage{cite}
\usepackage{lettrine}
\usepackage{amssymb,amsthm,epsfig,fancyhdr,bm,diagbox,multirow,booktabs,epstopdf,subfigure}
\usepackage{float,caption}
\usepackage[ruled,linesnumbered]{algorithm2e}
\SetKwComment{Comment}{ $\triangleright$ }{}
\def\tablestretch{1.5}

\begin{document}

\title{Autonomous Multi-Objective Optimization Using Large Language Model}

\author{
	Yuxiao~Huang,~
	Shenghao~Wu,~
	Wenjie~Zhang,~
	Jibin~Wu,~
	Liang~Feng,~
	and~Kay~Chen~Tan,~
\thanks{\textit{Corresponding author: Liang Feng}}

\thanks{Yuxiao Huang, Shenghao Wu are with the Department of Computing, The Hong Kong Polytechnic University, Hong Kong SAR. E-mail: \{yuxiao.huang, shenghao.wu\}@polyu.edu.hk.}
\thanks{Wenjie Zhang is with Department of Electrical and Elctronics Engineering, The Hong Kong Polytechnic University, Hong Kong SAR. E-mail: wenjie\_zhang@u.nus.edu}
\thanks{Jibin Wu and Kay Chen Tan are with the Department of Data Science and Artificial Intelligence, The Hong Kong Polytechnic University, Hong Kong SAR. E-mail: \{jibin.wu, kctan\}@polyu.edu.hk.}
\thanks{Liang Feng is with College of Computer Science, Chongqing University, Chongqing 400044, China. E-mail: \{liangf\}@cqu.edu.cn.}}



\maketitle

\begin{abstract}
Multi-objective optimization problems (MOPs) are ubiquitous in real-world applications, presenting a complex challenge of balancing multiple conflicting objectives. Traditional evolutionary algorithms (EAs), though effective, often rely on domain-specific expertise and iterative fine-tuning, hindering adaptability to unseen MOPs. In recent years, the advent of Large Language Models (LLMs) has revolutionized software engineering by enabling the autonomous generation and refinement of programs. Leveraging this breakthrough, we propose a new LLM-based framework that autonomously designs EA operators for solving MOPs. The proposed framework includes a robust testing module to refine the generated EA operator through error-driven dialogue with LLMs, a dynamic selection strategy along with informative prompting-based crossover and mutation to fit textual optimization pipeline. Our approach facilitates the design of EA operators without the extensive demands for expert intervention, thereby speeding up the innovation of EA operators. Empirical studies across various MOP categories validate the robustness and superior performance of our proposed framework.
\end{abstract}


\section{Introduction}

\lettrine[lines=2]{M}{ulti-objective} optimization is a crucial field of research that tackles the challenge of simultaneously optimizing two or more conflicting objectives in real-world scenarios\cite{gunantara2018review}. The primary goal of multi-objective optimization is to find a set of solutions, often known as the Pareto Optimal Set \cite{van1998evolutionary}, where no other solutions outperform all objectives. To solve the multi-objective optimization problems (MOPs), a lot of well-known evolutionary algorithms (EAs) have been developed in the literature, including non-dominated-sorting-based methods \cite{deb2002fast,zhou2021problem,deng2022enhanced} and decomposition-based techniques \cite{zhang2007moea,asafuddoula2014decomposition,dai2015new}. With the growing demand of multi-objective optimization in real-world applications \cite{tanabe2020easy}, extensive efforts have been conducted to improve the search performance of EAs \cite{HuangCentralized,XueSolutionTransfer,Fengmultiform}, with the goal of tackling the new multi-objective optimization challenges encountered. However, such hand-crafting pipeline for algorithm design necessitates a substantial amount of expert knowledge and numerous trials, which could constrain the innovation of EA in solving complex optimization problems.

Nowadays, owing to the impressive language processing capabilities, Large Language Models (LLMs) have made significant strides in software engineering tasks. Modern society increasingly leverages these powerful LLMs to address the demands of daily programming tasks \cite{belzner2023large,wu2024evolutionary,alto2023modern}. Particularly, Chen \textit{et al.} enabled LLMs to refine their programs autonomously \cite{chen2023teaching}, while Shypula \textit{et al.} proposed to enhance the efficiency of programs via few-shot prompting and chain-of-thought \cite{shypula2023learning}. Furthermore, efforts to create powerful programs using LLMs have explored iterative evolution, leading to new mathematical discoveries \cite{romera2024mathematical}, autonomous programming \cite{liventsev2023fully}, enhanced vehicle routing \cite{liu2023algorithm,liu2024example}, hybrid swarm intelligence \cite{pluhacek2023leveraging}, and dynamic heuristic adaptation \cite{ye2024reevo}. As these works demonstrate, LLMs are revolutionizing the field of software engineering through iterative program evolution, marking a new era in algorithmic design for optimization. 

While the existing methods of program evolution have showcased notable successes across various domains, it is worth noting that, they often adopt succinct prompts and traditional evolutionary strategies. Such design effectively develops pivotal task-specific functions with locked power of LLMs \cite{liu2024example,pluhacek2023leveraging,ye2024reevo}. However, when employed to construct intricate problem-solving systems for handling complex problems, such as MOPs, these methods often result in catastrophic outcomes---such as program crashes or endless loop---mainly due to inadequate error-handling mechanisms. Furthermore, they may encounter bottlenecks in program design arising from the absence of informative task descriptions and mismatched evolutionary strategies. Although these strategies perform well in numerical optimization tasks, they do not seamlessly align with the prompting-based textual optimization paradigm, which often lead to early stagnation of the search.

With the above in mind, to further exploit the power of LLM, we propose a new framework of EA operator design that utilizes LLMs to discover and refine EA operators tailored for diverse MOPs. Recognizing the susceptibility to errors and the prevalence of execution anomalies in sophisticated programs produced by LLMs, our framework incorporates a robust testing module, which is seamlessly integrated into the evolutionary progress. This module treats compilation and runtime errors as indicators, and then guides the refinement of designed operator through dialogues with LLMs using the collected errors. To safeguard against the risk of infinite execution loops---a known hazard in LLM-generated code---we launch this module as an independent process, governed by a strict execution time limit. Furthermore, we introduce a dynamic selection module, along with informative crossover and mutation that align with the prompting-based textual optimization paradigm, aims to enlarge the exploration capabilities of LLMs. Particularly, our proposed framework initiates with a diverse population of EA operators generated through LLMs. Throughout the evolutionary process, our proposed textual crossover and mutation techniques guide the LLM to craft advanced EA operators for various MOPs. These EA operators generated via LLMs may encompass various search strategies to effectively address MOPs. The contributions of this work can be outlined as follows:
\begin{itemize}
\item To handle the challenges posed by autonomous multi-objective optimization, a new EA operator evolution framework is devised, which facilitates the autonomous ideation and crafting of cutting-edge EA operators, thereby amplifying the search efficacy for solving MOPs.
\item In light of the intrinsic vulnerabilities and the frequently occurred errors in programs conceived by LLMs, a robust testing module is established, which can seamlessly refine the generated operator by leveraging the errors as a dialogue-based feedback with LLMs.
\item To enhance the performance of EA operator evolution, we introduce a dynamic selection strategy, along with informative crossover and mutation specifically tailored to the LLM-based textual optimization paradigm.
\item To validate the performance of our proposed framework, we have conducted extensive empirical studies, employing both continuous and combinatorial MOPs, against state-of-the-art human-engineered multi-objective optimization algorithms. The obtained results demonstrate the efficacy of EA operators generated via the proposed framework.
\end{itemize}

The remainder of this paper is structured as follows. Section \ref{sec:background} first provides a literature review of existing works on LLM-assisted program evolution for solving diverse problems, followed by the introduction of three distinct MOPs that serves as our case studies. Our proposed operator evolution framework by using LLM is presented in Section \ref{sec:method}. Furthermore, to assess the performance of the proposed framework, comprehensive empirical studies are conducted in Section \ref{sec:experiment}. Lastly, Section \ref{sec:conclusion} summarizes this work and discusses the directions to be explored in the future.

\section{Background} \label{sec:background}
This section begins with a literature review of the LLM-assisted program evolution methods. Subsequently, the mathematical definitions of continuous MOPs and combinatorial MOPs are given, which serve as the focal points of investigation in our study. Notably, the rationale behind the selection of these MOPs is to investigate the capability of LLM in devising innovative EA operators across varied gene encoding modes.

\subsection{Review of LLM-assisted Program Evolution}

The emergence of LLMs have marked a transformative phase in optimization \cite{wu2024evolutionary,huang2024exploring,huang2024multimodal}. Within the context of textual optimization, such as program evolution, LLMs empower the generation of code that addresses a wide range of programming requirements \cite{belzner2023large}. Early study of Lehman \textit{et al.} explored the potential of LLM trained to generate code in the context of genetic programming, which reveals that LLMs can significantly enhance the effectiveness of the programs \cite{lehman2023evolution}. Subsequently, Romera-Paredes \textit{et al.} utilized LLMs to make new mathematical discoveries via search novel programs iteratively, indicating the potential software engineering capability by the integration of LLMs and evolutionary computation \cite{romera2024mathematical}. Liventsev \textit{et al.} showcased the performance of fully autonomous programming using LLMs, suggesting a future where programs evolve without human guidance \cite{liventsev2023fully}. Liu \textit{et al.} proposed a LLM-based algorithm evolution framework to design novel programs for solving traveling salesman problems (TSPs) \cite{liu2023algorithm}. Furthermore, they extended their work to develop superior guided local search for combinatorial optimization \cite{liu2024example}. Pluhacek \textit{et al.} leveraged LLMs in meta-heuristic optimization to generate hybrid swarm intelligence, potentially outperforming traditional methods \cite{pluhacek2023leveraging}. Ye \textit{et al.} introduced LLMs as hyper-heuristics within a reflective evolution framework, allowing for dynamic adaptation \cite{ye2024reevo}. These advancements underscore the transformative role of LLMs in the algorithmic design for optimization tasks. 

Despite the promising advancements, the exploration of LLM-assisted program evolution in optimization remains embryonic. To date, no effort has been made to develop innovative EA operators using LLMs for MOPs. Our proposed methodology aims to bridge this gap, overcoming the intrinsic constraints of LLMs and crafting EA operators that deliver superior search performance across different MOPs.

\subsection{Continuous Multi-objective Optimization Problems}
Continuous Multi-objective Optimization Problems (CMOPs) constitute a pivotal category of optimization problems. These problems, characterized by continuous real numbers as decision variables, entail the optimization of multiple conflicting objectives. Considering a minimization CMOP, the mathematical definition can be given by:
\begin{equation} \label{eq:CMOP}
\begin{aligned}
\text{Minimize} \\
& \mathbf{f}(\mathbf{x}) = [f^1(\mathbf{x}), f^2(\mathbf{x}), \ldots, f^k(\mathbf{x})] \\
\text{Subject to} \\
& \mathbf{x} \in \mathcal{X} \\
\end{aligned}
\end{equation}
where $k$ represents the number of objectives, $\mathbf{x}$ denotes the decision variables and $\mathcal{X}$ specifies the feasible search space. Generally, the goal of CMOP is to find a set of Pareto-optimal solutions, $\mathcal{P}=\{\mathbf{x}_1, \mathbf{x}_2, \ldots\}, \mathbf{x} \in \mathcal{X}$, which cannot be improved in any objective without degrading another. The research value of CMOPs lies in their applicability to real-world scenarios, such as engineering design \cite{janga2007efficient}, finance \cite{chiam2008evolutionary}, and environmental management \cite{wang2019many}.

\subsection{Combinatorial Multi-objective Optimization Problems}
In contrast to the continuous MOPs which involve real-valued decision variables, combinatorial MOPs focus on discrete decision variables, which present challenges due to their discontinuity and lack of smoothness. In what follows, we introduce two distinct combinatorial MOPs, each characterized by unique gene encoding.
\subsubsection{Multi-objective Knapsack Problems}
MOKPs are combinatorial optimization problems that involve selecting a subset of items from a given set, subject to a capacity constraint \cite{bazgan2009solving}. Each item has multiple associated objectives (e.g., different types of profits). The goal is to find a Pareto-optimal set of items that maximizes the profits while respecting the capacity constraint. The mathematical definition of MOKPs can be expressed as follows. Given $n$ items with associated weights $\mathbf{w}_i$, $k$ types of profits $\mathbf{p}^{i}_j$ ($i=1,2,\cdots,k$), and a knapsack capacity $C$, the goal is to find a binary vector $\mathbf{x} = [\mathbf{x}_1, \mathbf{x}_2, \ldots]$ to maximize the profits given by:
\begin{equation} \label{eq:MOKP}
\begin{aligned}
\text{Maximize} \\
& \mathbf{f}(\mathbf{x}) = [f^1(\mathbf{x}), f^2(\mathbf{x}), \ldots, f^k(\mathbf{x})] \\
\text{Where} \\
& f^i(\mathbf{x}) = \sum_{j=1}^{n} \mathbf{p}^{i}_{j} \cdot \mathbf{x}_j \\
\text{Subject to} \\
& \sum_{j=1}^{n} \mathbf{w}_j \cdot \mathbf{x}_j \leq C, \\
& \mathbf{x}_j \in \{0,1\}, \quad j = 1,2,...,n
\end{aligned}
\end{equation}

MOKPs find applications in many real-world optimization problems, such as food order optimization \cite{khandekar2023dynamic}, engineering and operational research \cite{du2022mixed}, cargo loading and project selection \cite{song2022multiobjective}. The versatility of MOKPs in addressing these diverse challenges underscores their significance in strategic planning and optimization across various industries.

\subsubsection{Multi-objective Traveling Salesman Problems}

MOTSPs extend the traditional Traveling Salesman Problem (TSP) by incorporating several minimization objectives \cite{peng2009comparison}. In a MOTSP, a salesman is tasked with visiting each city in a set exactly once, while optimizing not just for the total travel distance, but also for additional objectives such as time, cost, or environmental impact. Mathematically, consider a set of $n$ cities and distinct $k$ types of pairwise distances $\mathbf{d}^i(u,v)$, where the $u$ and the $v$ specifies two different vertexes. The goal is to find a permutation $\pi$ of the $n$ cities that minimizes the $k$ traveling distances simultaneously, which can be given by:
\begin{equation} \label{eql:MOKP1}
\begin{aligned}
\text{Minimize} \\
& \mathbf{f}(\pi) = [f^1(\pi), f^2(\pi), \ldots, f^k(\pi)] \\
\text{Where} \\
& f^i(\pi) = \sum_{i=1}^{n-1} \mathbf{d}^i(\pi(i),\pi(i+1)) \\
\text{Subject to} \\
& \pi \in permutations(1,2,\ldots,n)
\end{aligned}
\end{equation}
Multi-objective Traveling Salesman Problems (MOTSPs) have practical implications in logistics, transportation, and network design. Particularly, MOTSPs streamline complex decision-making processes in areas such as efficient routing for delivery services \cite{khan2020multi}, sustainable urban development \cite{li2023solving}, and resource management in healthcare systems \cite{shuai2019effective}. The adaptability of MOTSPs to cater to multiple objectives simultaneously is what makes them invaluable in the pursuit of operational excellence and sustainability across diverse sectors.

\section{Proposed Method} \label{sec:method}

\begin{algorithm}[!b]
\caption{Pseudocode of the operator evolution.} \label{alg:total}
\KwIn{
	\\$\quad$ $G_{ev}$: Number of generations to evolve operators. \\
	\\$\quad$ $N_{ev}$: Number of operators in each population. \\
	\\$\quad$ $PBs$: Multi-objective problems.
}
\KwOut{
	\\$\quad$ $\mathcal{P}_{ev}$: The best evolutionary operators.
}
Obtain the initial operator population $\mathcal{P}_{ev}$ and the scores on $PBs$ via `Operator Initialization' (Alg. \ref{alg:initialization}).\\
\For{$gen \leftarrow 1$ to $G_{ev}$}
{
	\For{$i \leftarrow 1$ to $N_{ev}$}
	{
		Obtain selection probabilities of operators in $\mathcal{P}_{ev}$ based on their scores. \\
		Generate a random integer $N_s$, $1 < N_s \le N_{max}$\\
		Obtain a new operator $RC$ leveraging LLM and $N_s$ selected operators through `Operator Crossover' (Alg. \ref{alg:crossover}). \\
		Obtain a new operator $\hat{RC}$ leveraging LLM and the generated operator $RC$ through `Operator Mutation' (Alg. \ref{alg:mutation}). \\
		Obtain score of $\hat{RC}$ via parallel evaluation on $PBs$. \\
		$\mathcal{P}_{ev}=\mathcal{P}_{ev}\cup\{\hat{RC}\}$. \\
	}
	Update $\mathcal{P}_{ev}$ based on scores. \\
}
\end{algorithm}

\begin{figure*}[htbp]
\centering
\includegraphics[width=1.6\columnwidth]{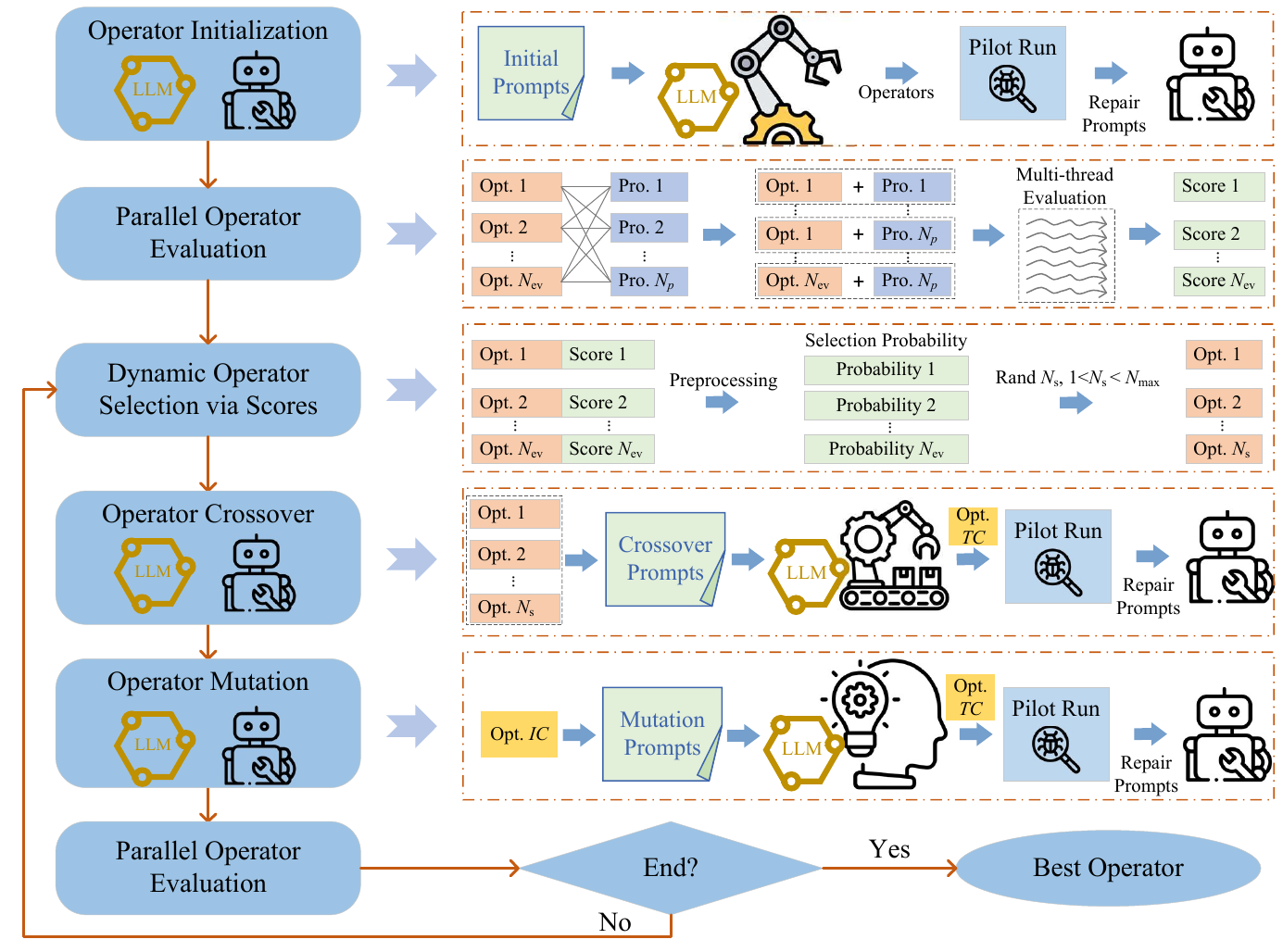}
\caption{Illustration of the proposed multi-objective algorithm evolution via LLM.}
\label{fig:workflow}
\end{figure*}

In this section, the proposed method for evolving EA operators via LLM is introduced. The workflow of our proposed paradigm has been depicted in Fig. \ref{fig:workflow}. In contrast to existing program evolution pipelines \cite{lehman2023evolution,liu2023algorithm,romera2024mathematical}, our proposed method integrates an LLM-assisted debugging process, encompassing Pilot Run \& Repair, into the core operator generation stages: ``Operator Initialization'', ``Operator Crossover'', and ``Operator Mutation''. Additionally, the ``Dynamic Operator Selection'' is tailored to complement the LLM-based crossover mechanism, facilitating the handling of textual data. 

As illustrated in Fig. \ref{fig:workflow} and outlined in Alg. \ref{alg:total}, the paradigm begins with the phase of ``Operator Initialization'', where an initial population of EA operators is generated from a meticulously crafted `Initial Prompts' (line 1 of Alg. \ref{alg:total}). These operators undergo ``Parallel Operator Evaluation'', receiving scores based on their efficacy across MOPs. These scores inform the ``Dynamic Operator Selection'' (lines 4-5), bringing diversity in ``Crossover Prompts'' by altering the number of selected parent operators, thus fostering the creation of innovative operators during ``Operator Crossover'' (line 6) and ``Operator Mutation'' (line 7). The newly generated operator obtained through mutation (i.e., $\hat{RC}$) is evaluated in parallel to obtain its score, which is then integrated into the population, followed by an elitist strategy-based update of the population (lines 8-11). The evolution of operators persists, driven by enhanced optimization performance, until the termination criteria are met.

The paradigm's iterative design guarantees ongoing refinement, cycling through operator deployment and assessment until an established endpoint is reached. With this framework, the most effective EA operators---demonstrating consistent superiority in solution generation---is identified. This LLM-aided strategy infuses program evolution with human-like deductive processes, bolstering the adaptability and resilience of multi-objective optimization methods. In what follows, the details of each component are introduced.

\subsection{Operator Initialization}
This section introduces the initialization of operators by prompting LLM. It is commonly acknowledged that well-crafted prompts are pivotal for the effective initialization of operators, which can significantly speed up the evolution of operators. In light of this, we advocate for the utilization of detailed task descriptions and stringent requirements to prompt the LLM, with the aim of engendering superior EA operators.

\begin{algorithm}[t]
\caption{Pseudocode of the operator initialization.} \label{alg:initialization}
\KwIn{
	\\$\quad$ $N_{ev}$: Number of operators in each population. \\
	\\$\quad$ $PBs$: Multi-objective problems.
}
\KwOut{
	\\$\quad$ $\mathcal{P}_{ev}$: The initialization population of operators.
}
\tcp{Operator Initialization}
$\mathcal{P}_{ev} \leftarrow \{\}$ \\
\For{$i \leftarrow 1$ to $N_{ev}$}
{
	$RC \leftarrow None$ \\
	\While{$RC$ is $None$}
	{
		Obtain an operator $TC$ via prompting LLM. \\
		Obtain new $RC$ via Alg. \ref{alg:repair}. \\
	}
	Obtain score of $RC$ via parallel evaluation on $PBs$. \\
	$\mathcal{P}_{ev}=\mathcal{P}_{ev}\cup\{RC\}$ \\
}
\end{algorithm}

The prompts for initialization are presented in the Fig. \ref{prompt:initialization}, titled by ``INITIALIZATION''. As depicted, the prompts can be briefly divided into three parts, i.e., `System', `Description of Task' and `Requirements', respectively. The `System' gives the initial prompts for LLM with a brief description for the task. The `Description of Task' outlines the details of the task for the LLM, while the `Requirements' provides the specific requirements on the task. Within these prompts, the placeholder \#PROBLEM\# is a user-defined descriptor for MOPs; the \#PROBLEM\_DESC\# presents an informative introduction to the MOP with optimization objectives, which aids crafting bespoke search operator for specific MOPs; and the \#FORMAT\# delineates the operator's specifics, encompassing the function name, input parameters, output and their annotations. Details of these placeholders have been given in the Supplementary Materials. Employing these comprehensive prompts enables the LLM to generate EA operators that may achieve higher score across MOPs, as delineated in Alg. \ref{alg:initialization}. To further validate the operator produced by the LLM engine, it will undergo Pilot Run \& Repair as discussed in Section \ref{sec:repair}, specifically lines 3-7 of Alg. \ref{alg:initialization}. The generation process continues until an operator successfully passes the validation tests or a rectified operator is produced through the repair process outlined in Alg. \ref{alg:repair}.

\begin{center}
\fcolorbox{black}{cyan!20}{\parbox{0.9\linewidth}{
\begin{center}
\textcolor[rgb]{0.0,0.0,0.0}{\textbf{INITIALIZATION}}\\
\end{center}
\textcolor[rgb]{0.5,0.6,0.7}{\textbf{
System}:\\
You are an expert in designing intelligent evolutionary search strategies that can solve \#PROBLEM\# efficiently and effectively. \#PROBLEM\_DESC\# \\
}
\textcolor{blue}{\textbf{
Description of Task}:\\
Your task is to evolve a superior evolutionary operator with Python for tackling \#PROBLEM\#, with the goal of achieving top search performance across \#PROBLEM\#. You have to provide me the Python code with a single function namely `next\_generation' following the format and the requirements given below, which are matched with their functionalities: \\
\#FORMAT\# \\
}
\textcolor{magenta}{\textbf{
Requirements}:\\
You have to return me a single function namely `next\_generation', keep the format of input and the format of output unchanged, and provide concise descriptions in the annotation. \\
Please return me an XML text using the following format:\\
$<$next\_generation$>$\\
...\\
$<$/next\_generation$>$\\
where `...' gives only the entire code without any additional information. To enable direct compilation for the code given in `...', please don't provide any other text except the single Python function namely `next\_generation' with its annotation.\\
No Explanation Needed!! \\
}
}}
\captionof{figure}{Prompt design for `Operator Initialization'}
\label{prompt:initialization}
\end{center}

\subsection{Dynamic Operator Selection via Scores}
This section introduces the proposed selection strategy tailored for LLM-assisted operator evolution. In traditional evolutionary algorithms \cite{back1993overview, bartz2014evolutionary, he2016many}, this component is typically responsible for selecting a predetermined number of parents (commonly two) for subsequent operations such as crossover. In contrast, generative pre-trained transformer (GPT) models \cite{zhu2022generative, luo2022biogpt} may yield similar responses (i.e., programs) when provided with a detailed and specific prompt---effectively replicating the same operator. This can lead to a stagnation in the operator evolution progress, as the newly generated operator, when underperforming compared to the existing operators within the operator population, may cause the process to become ensnared in local optima. 

Considering the aforementioned issue, this study introduces a dynamic selection mechanism to facilitate the generation of novel operators during operator crossover. Specifically, as illustrated in lines 4-5 of Alg. \ref{alg:total}, the selection probabilities for the individual operators within $\mathcal{P}_{ev}$ are determined based on their performance scores on MOPs. A higher score signifies enhanced search capability and, consequently, an increased likelihood of selection. The selection probability for each operator is computed as follows:
\begin{equation} \label{eql:prob_calc}
P_s(i) = \frac{\exp(score_i)}{\sum_{k=1}^{N_{ev}}\exp(score_k)}
\end{equation}
where $N_{ev}$ denotes the number of operators within the population. Additionally, a random integer $N_s$, where $1 < N_s \le N_{max}$, is generated to designate the count of parent operators chosen for the operator crossover.

\subsection{Operator Crossover}
Utilizing the dynamically selected parent operators, this study performs informative operator crossovers by engaging a LLM with these operators. The LLM, serving as an intelligent engine, is anticipated to synthesize innovative operators that surpass the parent operators in terms of scores. The prompts for the LLM are structured as Fig. \ref{prompt:crossover}, named by ``CROSSOVER'':
\begin{center}
\fcolorbox{black}{cyan!20}{\parbox{0.9\linewidth}{
\begin{center}
\textcolor[rgb]{0.0,0.0,0.0}{\textbf{CROSSOVER}}\\
\end{center}
\textcolor{blue}{\textbf{
Description of Task}:\\
I will showcase several evaluated `next\_generation' functions in XML format, with their scores obtained on the \#PROBLEM\#. Your task is to conceive an advanced function with the same input/output formats, termed `next\_generation', that is inspired by the evaluated cases.\\
\#FORMAT\#\\
Below, you will find the \#$N_s$\# evaluated `next\_generation' functions in XML texts, each accompanied by its corresponding score.\\
\#SELECTED\_OPERATORS\#\\
}
\textcolor{magenta}{\textbf{
Requirements}:\\
Kindly devise an innovative `next\_generation' method with XML that retains the identical input/output structure. This method should be crafted through a meticulous analysis of the shared characteristics among high-performing algorithms.\\
No Explanation Needed!! \\
}
}}
\captionof{figure}{Prompt design for `Operator Crossover'}
\label{prompt:crossover}
\end{center}

The placeholders \#PROBLEM\#, \#FORMAT\#, \#$N_s$\#, and \#SELECTED\_OPERATORS\# denote the MOP name, the coding format of the operator, the quantity of parent operators selected, and the XML code snippets of the parent operators with their scores, respectively. The dynamic nature of these prompts is likely to facilitate the generation of varied operators by the LLM, as outlined in line 3 of Alg. \ref{alg:crossover}. However, as outlined in line 4 of Alg. \ref{alg:total}, the resultant operator $TC$ requires additional validation or refinement to obtain $RC$, akin to the proposed operator initialization.

\begin{algorithm}[t]
\caption{Pseudocode of the operator crossover.} \label{alg:crossover}
\KwIn{
	\\$\quad$ $\mathcal{\hat{P}}_{ev}$: Selected operators via dynamic operator selection. \\
}
\KwOut{
	\\$\quad$ $RC$: Validated or repaired operator. \\
}
\tcp{Operator Crossover}
$RC \leftarrow None$ \\
\While{$RC$ is $None$}
{
	Obtain a new operator $TC$ via prompting LLM with $\mathcal{\hat{P}}_{ev}$. \\
	Obtain new $RC$ via Alg. \ref{alg:repair}. \\
}
\end{algorithm}

\subsection{Operator Mutation}
This section introduces the process of operator mutation, facilitated through the utilization of a LLM. The fundamental goal of this mutation component is to implement strategic modifications to the operators derived from the crossover phase. As illustrated in lines 1-7 of Alg. \ref{alg:mutation}, these alterations occur with a predefined probability, i.e., $1/N_{ev}$, where $N_{ev}$ denotes the total number of operators within the population $\mathcal{P}$.
\begin{algorithm}[t]
\caption{Pseudocode of the operator mutation.} \label{alg:mutation}
\KwIn{
	\\$\quad$ $IC$: Input operator gained via operator crossover. \\
}
\KwOut{
	\\$\quad$ $\mathcal{P}_{ev}$: The best evolutionary operators.
}
\tcp{Operator Mutation}
\If{$rand<\frac{1}{N_{ev}}$}
{
	$RC \leftarrow None$ \\
	\While{$RC$ is $None$}
	{
		Obtain a new operator $TC$ via prompting LLM with $IC$. \\
		Obtain new $RC$ via Alg. \ref{alg:repair}. \\
	}
}
\Else{
	$RC \leftarrow IC$ \\
}
\end{algorithm}
Given an input operator, i.e., $IC$, we intend to alter the code slightly with the prompts given by Fig. \ref{prompt:mutation}, where the placeholder \#OPERATOR\# represents the textual data of $IC$. Once gained a modified operator  by leveraging the LLM, the operator (i.e., $TC$) will be validated or repaired through Alg. \ref{alg:repair} akin to `Operator Crossover' (line 5 of Alg. \ref{alg:mutation}).

The mutation mechanism is designed to be adaptive, allowing for the fine-tuning of operators based on the evolving demands of the MOPs at hand. Such component mitigates the risk of premature convergence and maintains a healthy diversity among operators, which is crucial for navigating complex problem landscapes.

\begin{center}
\fcolorbox{black}{cyan!20}{\parbox{0.9\linewidth}{
\begin{center}
\textcolor[rgb]{0.0,0.0,0.0}{\textbf{MUTATION}}\\
\end{center}
\textcolor{blue}{\textbf{
Description of Task}:\\
I will introduce an evolutionary search function titled `next\_generation' in XML format. Your task is to meticulously refine this function and propose a novel one that may obtain superior search performance on \#PROBLEM\#, ensuring the input/output formats, function name, and core functionality remain unaltered.\\
\#FORMAT\#\\
The original function is given by:\\
$<$next\_generation$>$\#OPERATOR\#$<$/next\_generation$>$\\
}
\textcolor{magenta}{\textbf{
Requirements}:\\
Please return me an innovative `next\_generation' operator with the same XML format. No Explanation Needed!! \\
}
}}
\captionof{figure}{Prompt design for `Operator Mutation'}
\label{prompt:mutation}
\end{center}

\subsection{Pilot Run \& Repair} \label{sec:repair}
Since the operators generated via LLMs may encounter various errors during compilation and execution, which can lead to disastrous outcomes in operator evolution, it is of great significance to design a component that ensures the quality of operator produced by LLMs. With this goal in mind, we introduce a Pilot Run \& Repair component in this section, seamlessly integrated into the program's evolution process. This component treats compilation and runtime errors as indicators and engages in dialogues with LLMs to guide operator quality improvement.

\begin{center}
\fcolorbox{black}{cyan!20}{\parbox{0.9\linewidth}{
\begin{center}
\textcolor[rgb]{0.0,0.0,0.0}{\textbf{REPAIR}}\\
\end{center}
\textcolor{blue}{\textbf{
Description of Task}:\\
The code you provided for me cannot pass my demo test on \#PROBLEM\#. The error is given by: \#ERROR\#. Can you correct the code according to the errors?\\
}
\textcolor{magenta}{\textbf{
Requirements}:\\
Please return me a refined `next\_generation' with the same XML format, i.e., \\$<$next\_generation$>$...$<$/next\_generation$>$\\, where the `...' represents the code snippet. \\
No Explanation Needed!! \\
}
}}
\captionof{figure}{Prompt design for repairing operators}
\label{prompt:repair}
\end{center}

Alg. \ref{alg:repair} outlines the workflow of the proposed component. It assesses the quality of the tested operator (referred to as $TC$) on the toy problems (given by $PBs$) within a specified budgeted running time ($MaxT$). If necessary, it proceeds to repair the operator by identifying errors and engaging in dialogues with LLMs. The procedure initiates a while loop with a maximum repair time ($N_{trail}$). Within the loop, as observed in lines 3-4, an independent process is launched to execute $RC$ on $PBs$, adhering to the time budget of $MaxT$. The process provides its running state and any encountered errors during execution (line 5). Notably, the error may be empty due to unknown issues (such as an endless loop). Subsequently, if $state=True$, it signifies that the tested operator successfully passed the pilot run and does not require further repair through LLM dialogues (lines 6-8). Conversely, when $state=False$ and $error$ is not empty, it indicates that errors occurred during testing and can be leveraged for code repair using LLMs (lines 9-11). The code repair prompts are illustrated in Fig. \ref{prompt:repair}, termed as `REPAIR', where \#PROBLEM\# denotes the name of the test MOPs, and \#ERROR\# encapsulates the descriptions collected by the error handler. Otherwise, if no specific issues are identified, $None$ is returned to denote default testing failure (lines 12-14). The repair process continues until the specified maximum repair time ($N_{trail}$) is met, ultimately returning $None$.

\begin{algorithm}[t]
\caption{Pseudocode of Pilot Run \& Repair.} \label{alg:repair}
\KwIn{
	\\$\quad$ $N_{trail}$: Number of trails for code repair. \\
	\\$\quad$ $TC$: The tested operator. \\
	\\$\quad$ $PBs$: The testing multi-objective problems. \\
	\\$\quad$ $MaxT$: The maximum running time. 
}
\KwOut{
	\\$\quad$ $RC$: Repaired operator.
}
$iter \leftarrow 0$ \\
\While{$iter < N_{trail}$}
{
	$RC \leftarrow TC$ \\
	Start an independent process to run $RC$ on $PBs$ under a limited running time $MaxT$. \\
	Obtain the running $state$ and $error$ occurred during the testing run. \\
	\If{$state$ is $True$}
	{
		Return $RC$ \\
	}
	\ElseIf{$state$ is $False$ and $error$ is not empty}
	{
		Obtain new $TC$ based on communication with LLM using the informative $error$ collected.\\
	}
	\Else
	{
		Return $None$ \\
	}
    $iter \leftarrow iter + 1$ \\
}
Return $None$ \\
\end{algorithm}


\subsection{Parallel Operator Evaluation}
Upon successful completion of the `Pilot Run' by the operators generated through LLM, a parallel evaluation is conducted to measure their efficacy on the MOPs in terms of a normalized score. As depicted in Fig. \ref{fig:workflow}, each operator, denoted as `Opt. $*$', is systematically paired with a validation problem, referred to `Pro. $*$'). Following this, the pairs are independently processed in multiple threads, each yielding an operator's performance score for a specific MOP. It is worth noting that, the normalized score for the $i$-th operator, aggregated across the MOPs, is calculated as follows:
\begin{equation} \label{eql:prob_calc}
score_{i}=MEAN(PS)-STD(PS)
\end{equation}
Here, `$MEAN$' and `$STD$' denote the mean and standard deviation of the collected scores, respective. $PS$ specifies the scores obtained on the MOPs. By maximizing the normalized $score$ through LLM-assisted operator evolution, we aim to derive an operator that not only excels consistently across MOPs but also demonstrates robust competitiveness across diverse MOPs.

\section{Experimental Study} \label{sec:experiment}
In this section, we conduct comprehensive empirical studies to rigorously assess the efficacy of our proposed LLM-assisted operator evolution framework within the domain of multi-objective optimization. This section begins by delineating the experimental settings of validation and testing MOPs, the performance metrics across different MOP categories, and the setup of compared multi-objective algorithms. Subsequently, the results obtained on COMPs, MOKPs and MOTSPs are presented and discussed to validate the search performance of the proposed automatic operator evolution paradigm, which is denoted by LLMOPT.

\subsection{Experimental Settings}
We select one continuous MOP and two combinatorial MOPs with very different gene encoding modes to validate the performance of our proposed framework: the CMOP with continuous variables, the MOKP with binary variables, and the MOTSP with permutation-based variables. Each MOP category is further subdivided into two distinct sets: a validation group and a testing group. The validation group consists of smaller-scale problem instances that are utilized during the evolutionary progression of operators in LLMOPT. On the other hand, the testing group comprises larger-scale problem instances that serve to assess the performance of the best operator optimized through our proposed LLMOPT.

For the CMOPs, we select `ZDT1' to `ZDT6' (two objectives) and `DTLZ1' to `DTLZ7' (three objectives) \cite{tian2018sampling} with default configurations (with decision variables ranging from 10 to 30) in \cite{blank2020pymoo} as the validation group, while their counterparts with 50 decision variables serve as the testing group. The performance score for each CMOP instance, denoted by $PS_{i}$, is calculated via $PS_{i}=1-IGD_{i}$, where $IGD_i$ denotes the Inverted Generational Distance (IGD) value \cite{sun2018igd} obtained for the $i$-th problem instance. A lower IGD value signifies a higher score and indicates a more effective search performance by the operator. In the case of the MOKP, we have randomly generated 10 problem instances with item counts ranging from 50 to 200 for the validation group, and 10 problem instances with item counts ranging from 100 to 200 for the testing group. The score for each MOKP instance is given by: $PS_{i}=1-HV_{i}$, where $HV_{i}$ denotes the Hypervolume (normalized between 0 and 1) \cite{bradstreet2011hypervolume} achieved on the $i$-th problem instance. Since the MOKP is a maximization problem, a lower Hypervolume implies a higher score and more favorable outcomes. Furthermore, for the MOTSP, we have generated 20 problem instances with 30 vertices for the validation group, and 10 problem instances with vertices ranging from 100 to 200 for the testing group. The score for each MOTSP instance is determined by: $PS_{i}=HV_{i}$. As the MOTSP is a minimization problem, a higher normalized Hypervolume indicates a higher score and a more proficient search capability.


\begin{table*}[htbp]
\centering
\caption{Configurations of MOEAD, CTAEA, RVEA, NSGA2, AGEMOEA and SMSEMOA for solving different categories of MOPs.}
\setlength{\tabcolsep}{4pt} 
\renewcommand{\arraystretch}{\tablestretch} 
\begin{tabular}{cc|cccccc}
\hline
\multicolumn{2}{l|}{\multirow{2}{*}{}} & \multicolumn{3}{c}{Decomposition-based}                                                                            & \multicolumn{3}{c}{Non-dominated Sorting-based}                                                                    \\
\multicolumn{2}{l|}{}                  & MOEAD                                & CTAEA                                & RVEA                                 & NSGA2                                & AGEMOEA                              & SMSEMOA                              \\ \hline
\multicolumn{2}{c|}{Population Size}         & 100                                  & 100                                  & 100                                  & 100                                  & 100                                  & 100                                  \\
\multicolumn{2}{c|}{Generation}        & 200                                  & 200                                  & 200                                  & 200                                  & 200                                  & 200                                  \\ \hline
\multirow{3}{*}{Crossover}   & CMOP    & SBX                                  & SBX                                  & SBX                                  & SBX                                  & SBX                                  & SBX                                  \\
								& MOKP    & ---                                  & TwoPointCrossover                    & TwoPointCrossover                    & TwoPointCrossover                    & TwoPointCrossover                    & TwoPointCrossover                    \\
								& MOTSP   & OrderCrossover                       & OrderCrossover                       & OrderCrossover                       & OrderCrossover                       & OrderCrossover                       & OrderCrossover                       \\ \hline
\multirow{3}{*}{Mutation}    & CMOP    & PM                                   & PM                                   & PM                                   & PM                                   & PM                                   & PM                                   \\
								& MOKP    & ---                                  & BitflipMutation                      & BitflipMutation                      & BitflipMutation                      & BitflipMutation                      & BitflipMutation                      \\
								& MOTSP   & InversionMutation                    & InversionMutation                    & InversionMutation                    & InversionMutation                    & InversionMutation                    & InversionMutation                    \\ \hline
Repair                       & MOKP    & \multicolumn{1}{l}{RandWeightRepair} & \multicolumn{1}{l}{RandWeightRepair} & \multicolumn{1}{l}{RandWeightRepair} & \multicolumn{1}{l}{RandWeightRepair} & \multicolumn{1}{l}{RandWeightRepair} & \multicolumn{1}{l}{RandWeightRepair} \\ \hline
\end{tabular}
\label{tab:config}
\end{table*}

\begin{figure*}[htbp]
\centering
\subfigure[Validation CMOPs]{
\includegraphics[width=0.64\columnwidth]{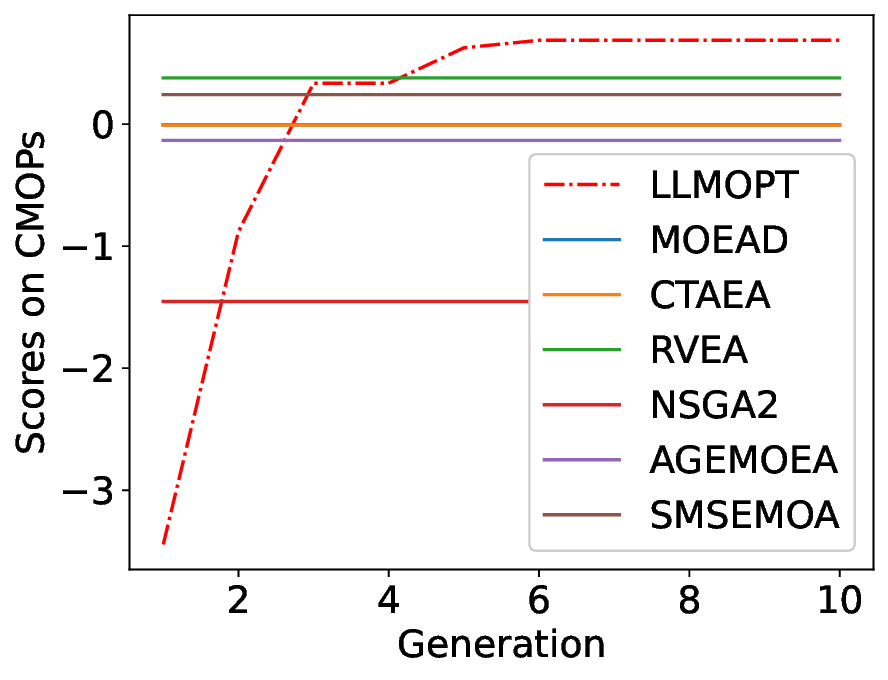}}
\subfigure[Validation MOKPs]{
\includegraphics[width=0.67\columnwidth]{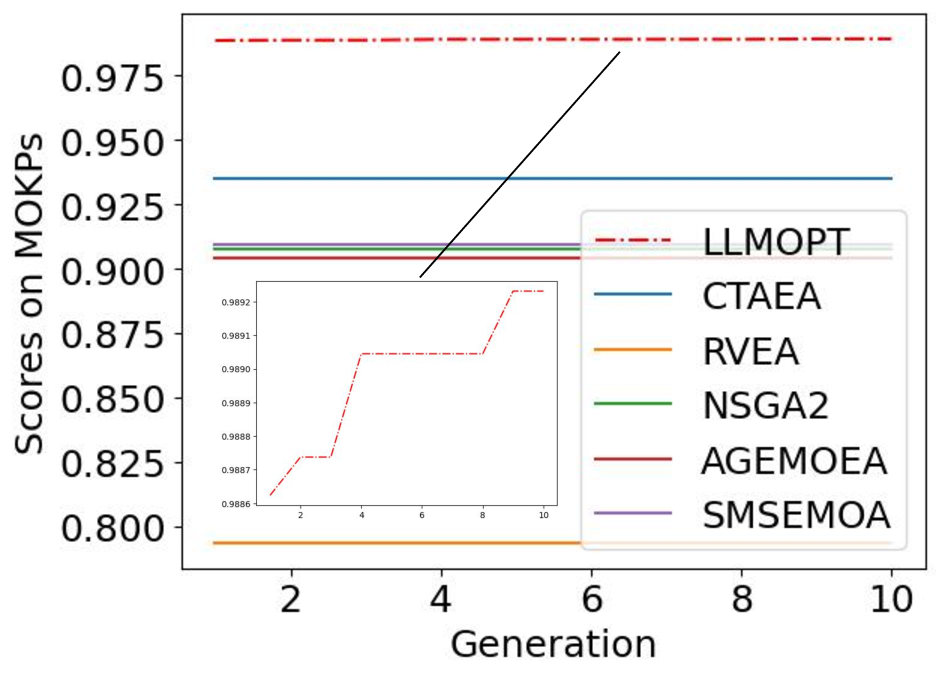}}
\subfigure[Validation MOTSPs]{
\includegraphics[width=0.65\columnwidth]{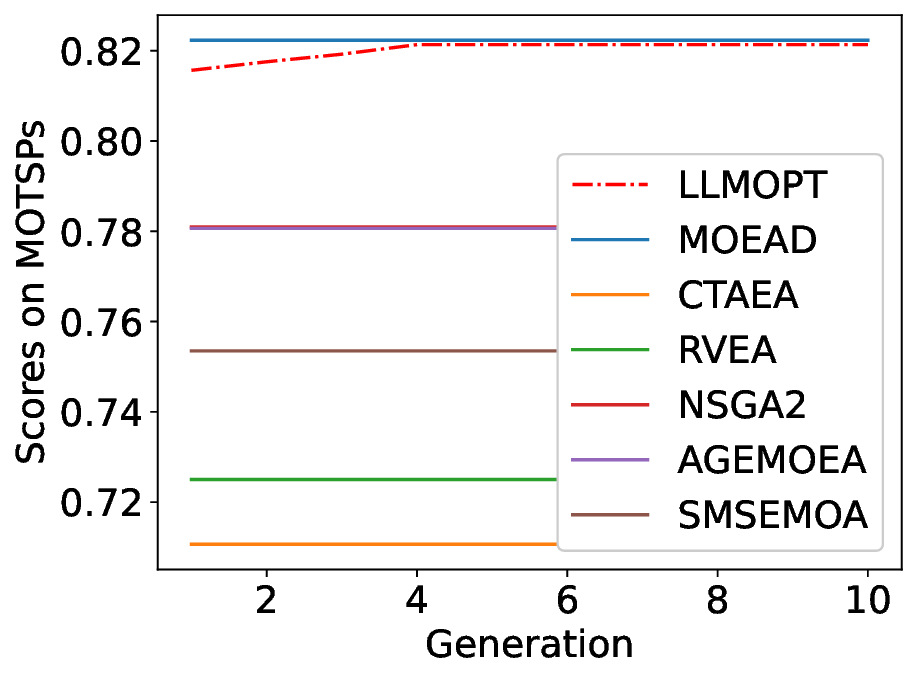}}
\caption{The convergence curves of the proposed operator evolution framework on different types of MOPs.}
\label{fig:LLMEvolve}
\end{figure*}

To demonstrate the search performance of the EA operator optimized via the proposed LLMOPT, we adopt six multi-objective optimization algorithms with suitable evolutionary operators for handling the three MOP categories based on \cite{blank2020pymoo}, namely MOEAD \cite{zhang2007moea}, CTAEA \cite{li2018two}, RVEA \cite{cheng2016reference}, NSGA2 \cite{deb2002fast}, AGEMOEA \cite{panichella2019adaptive}, SMSEMOA \cite{beume2007sms}. The configurations of these methods are detailed in Table \ref{tab:config}, ensuring a fair comparison by maintaining the same population size and the maximum number of generations across all algorithms (also configured in the operator obtained through LLMOPT). As can be observed, for CMOPs, we employ the `Simulated Binary Crossover (SBX)' \cite{deb1995simulated} and `Polynomial Mutation (PM)' \cite{deb2007self} operators; MOKPs are addressed using `Two-Point Crossover' \cite{ouerfelli2013genetic} and `Bit-Flip Mutation' \cite{garnier1999rigorous}; while MOTSPs are optimized with `Order Crossover' \cite{ono1996genetic} and `Inversion Mutation' \cite{schmid1983genetic}. Notably, the MOEAD is not applied to MOKPs due to its incompatibility with the knapsack capacity constraints. Additionally, for the efficient resolution of MOKPs, the commonly adopted `RandWeightRepair' strategy is employed, wherein items are randomly removed to meet the capacity constraints. Each multi-objective algorithm under comparison, as well as the optimized operator, is subjected to 10 independent runs for every MOP instance to ensure robust statistical analysis. The hyper-parameters specific to the evolution of the operator within our framework are as follows:
\begin{itemize}
\item Population size for the proposed framework: $N_{ev}=10$.
\item Maximum generations for the proposed framework: $G_{ev}=10$.
\item Maximum number of selected parents: $N_{max}=N_{ev}/2$.
\item Language model employed: gpt-4-1106-preview.
\item Temperature setting for the language model: $0.5$.
\item Maximum running time for Pilot Run: $MaxT=2000s$.
\item Number of trials for code repair: $N_{trial}=2$.
\end{itemize}
The temperature is set at $0.5$ to strike an optimal balance between exploitation and exploration within the operator evolution framework. Other configuration of the temperature can also be applied accordingly. The maximum running time $MaxT$ and the number of trials $N_{trial}=2$ for code repair are configured empirically.

\subsection{Results and Discussions}

In this section, we first evaluate the efficacy of the proposed operator evolution framework, i.e., LLMOPT, on the previously discussed validation MOPs. Figure \ref{fig:LLMEvolve} depicts the convergence trajectories of the proposed LLMOPT across these MOPs. The X-axis represents the evolutionary progress of the operators, while the Y-axis indicates the scores achieved by the best-performing operator in the current population. As shown in Fig. \ref{fig:LLMEvolve} (a), the performance of EA operator generated at the early stage is relatively low compared to other multi-objective algorithms due to the black-box nature of CMOPs, which provide limited problem information. Despite initial trials, the generated operators for CMOPs were unsatisfactory. However, the proposed LLMOPT allows the LLM to iteratively refine EA operators through feedback, showcasing the advantage of our approach over one-off code generation techniques. At last, the optimized EA operator obtained the best performance in terms of the scores on validation CMOPs. Sub-figure (b) reveals that the EA operator generated through LLMOPT outperforms competing algorithms from the start in the MOKP domain, which leverages on the rich problem-specific properties within the informative prompts. Notably, the optimization capabilities of these EA operators continue to improve as the evolution progresses. Similarly, sub-figure (c) shows that LLMOPT starts with a high initial score against CTAEA, RVEA, NSGA2, AGEMOEA and SMSEMOA. By iteratively improving the performance via the proposed LLMOPT, the optimized EA operator ultimately achieved competitive performance against MOEAD, which obtained the best performance on validation MOTSPs. Despite this, MOEAD shows weaker results in CMOPs, underscoring the competitive edge of operators optimized via our proposed method across diverse MOP categories.

\begin{table*}[htbp]
\centering
\caption{Averaged IGD values obtained by MOEAD, CTAEA, RVEA, NSGA2, AGEMOEA, SMSEMOA, and the best operator optimized via the proposed LLMOPT on testing CMOP instances over 10 independent runs. Superior averaged results on each instance are highlighted in bold font.}
\setlength{\tabcolsep}{8pt} 
\renewcommand{\arraystretch}{\tablestretch} 
\begin{tabular}{lccccccc}
\toprule
Problem & LLMOPT & MOEAD & CTAEA & RVEA & NSGA2 & AGEMOEA & SMSEMOA \\
\midrule
zdt1 & 1.133e-02 & 2.575e-02 & 2.313e-02 & 7.249e-02 & 9.998e-03 & 1.098e-02 & \textbf{6.569e-03} \\ 
zdt2 & 3.996e-02 & 1.979e-01 & 6.199e-02 & 1.079e-01 & \textbf{1.320e-02} & 2.779e-02 & 5.899e-02 \\ 
zdt3 & 1.947e-02 & 3.069e-02 & 3.211e-02 & 1.120e-01 & 8.365e-03 & 7.526e-03 & \textbf{6.567e-03} \\ 
zdt4 & 3.180e+01 & \textbf{1.093e+01} & 3.199e+01 & 5.346e+01 & 3.673e+01 & 3.140e+01 & 3.521e+01 \\ 
zdt5 & 1.587e-01 & 2.055e-01 & 1.836e-01 & 1.749e-01 & \textbf{1.266e-01} & 1.424e-01 & 1.462e-01 \\ 
zdt6 & \textbf{1.984e-02} & 3.116e-01 & 2.445e+00 & 1.870e+00 & 1.321e+00 & 1.052e+00 & 1.487e+00 \\ 
dtlz1 & \textbf{6.519e+01} & 1.011e+02 & 1.179e+02 & 8.428e+01 & 2.503e+02 & 1.861e+02 & 1.676e+02 \\ 
dtlz2 & 7.758e-02 & 7.506e-02 & 6.032e-02 & \textbf{5.857e-02} & 8.110e-02 & 6.666e-02 & 6.662e-02 \\ 
dtlz3 & \textbf{2.054e+02} & 3.129e+02 & 4.434e+02 & 3.113e+02 & 5.438e+02 & 4.360e+02 & 5.571e+02 \\ 
dtlz4 & 3.982e-01 & 6.064e-01 & \textbf{5.819e-02} & 6.359e-02 & 8.066e-02 & 4.409e-01 & 2.536e-01 \\ 
dtlz5 & 1.564e-02 & 3.028e-02 & 1.065e-01 & 1.264e-01 & 1.111e-02 & 1.443e-02 & \textbf{6.789e-03} \\ 
dtlz6 & \textbf{2.318e-01} & 3.055e+01 & 2.506e+01 & 2.327e+01 & 2.658e+01 & 2.801e+01 & 1.850e+01 \\ 
dtlz7 & 1.169e-01 & 2.041e-01 & 1.440e-01 & 3.143e-01 & 1.238e-01 & 1.272e-01 & \textbf{9.400e-02} \\ 
Averaged IGD & \textbf{2.335e+01} & 3.516e+01 & 4.780e+01 & 3.655e+01 & 6.609e+01 & 5.257e+01 & 6.004e+01 \\ 
\bottomrule
\end{tabular}
\label{tab:CMOP}
\end{table*}

\begin{figure*}[htbp]
\centering
\subfigure[ZDT-6]{
\includegraphics[width=0.415\columnwidth]{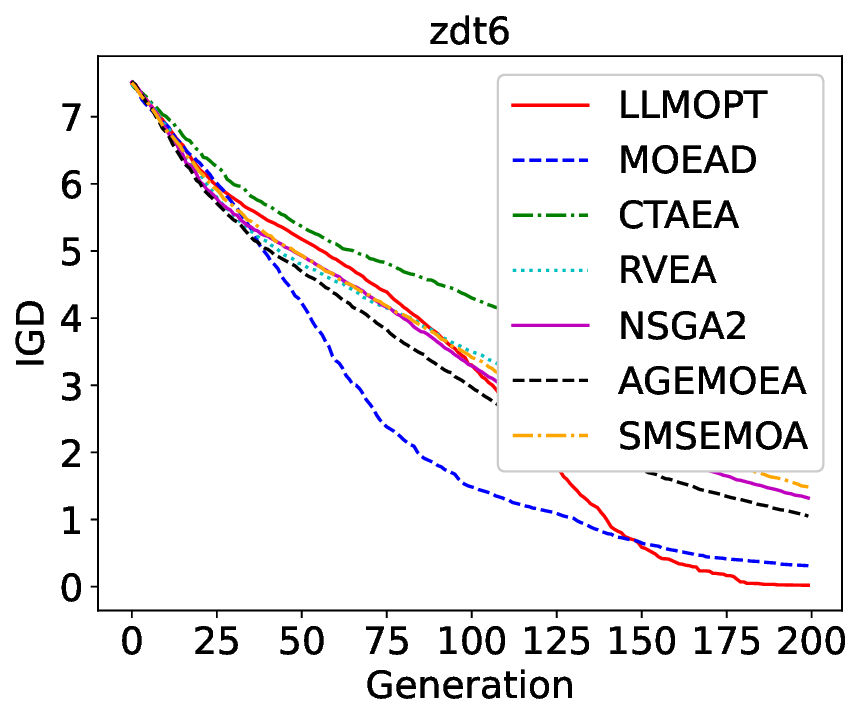}}
\subfigure[DTLZ-1]{
\includegraphics[width=0.45\columnwidth]{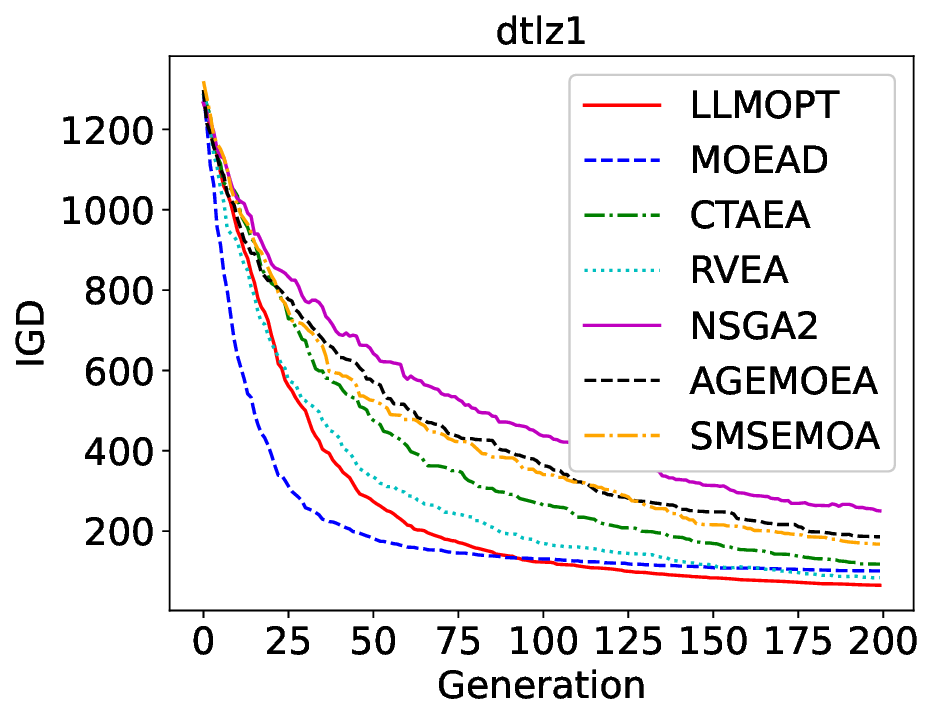}}
\subfigure[DTLZ-3]{
\includegraphics[width=0.455\columnwidth]{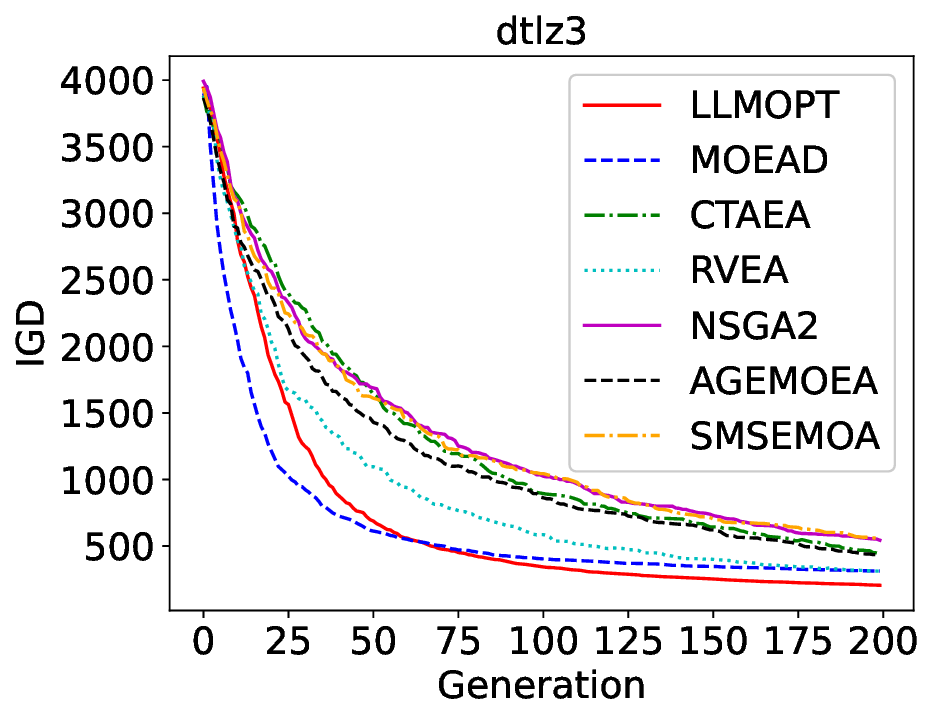}}
\subfigure[DTLZ-6]{
\includegraphics[width=0.43\columnwidth]{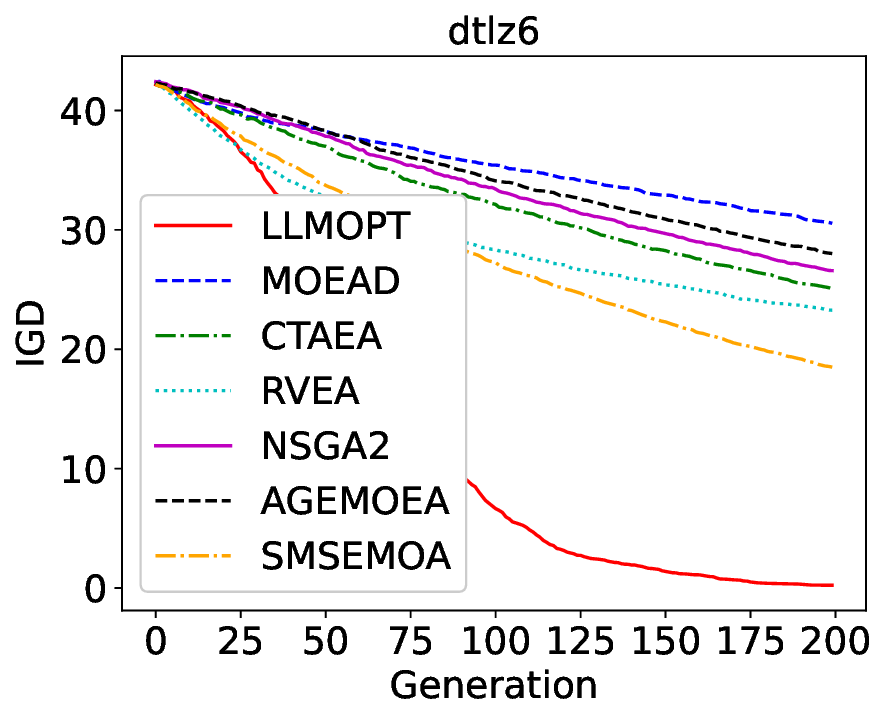}}
\caption{Convergence curves obtained by MOEAD, CTAEA, RVEA, NSGA2, AGEMOEA, SMSEMOA, and the best EA operator generated by the proposed LLMOPT on the representative CMOPs.}
\label{fig:CMOP}
\end{figure*}

To delve deeper into the search capabilities of the best operator optimized via the proposed LLMOPT, we present a detailed analysis of the results obtained on the testing CMOPs, MOKPs and MOTSPs.

\subsubsection{Evaluation on CMOPs}

Table \ref{tab:CMOP} presents the averaged IGD values achieved by MOEAD, CTAEA, RVEA, NSGA-II, AGEMOEA, SMSEMOA, and the best operator evolved by LLMOPT across testing CMOP instances, based on 10 independent runs. The `Average IGD' row reflects the mean IGD value across all evaluated problems. It is evident that the best EA operator generated by LLMOPT secures competitive standings when juxtaposed with existing hand-crafted multi-objective optimization methods. Notably, it excels in more challenging scenarios (from dtlz1 to dtlz7), outperforming all compared methods in three instances. This success can be attributed to the scoring mechanism defined in Eq. \ref{eql:prob_calc}, which compels the LLM-assisted search engine to harmonize search efficacy across all problems, thereby significantly diminishing large IGD values in complex problems like `DTLZ-1' and `DTLZ-3'. This feedback loop culminates in the lowest average IGD for the testing CMOPs. Additionally, Fig.\ref{fig:CMOP} illustrates the convergence curves for each algorithm on representative CMOP cases. The operator optimized by LLMOPT demonstrates rapid convergence---or at least competitive convergence rates---in most instances, such as `ZDT-6', `DTLZ1', `DTLZ-3', and `DTLZ-6'. The marked reduction in IGD value for `DTLZ-6' further underscores the performance of the operator obtained by LLMOPT in contrast to other methods.

\subsubsection{Evaluation on MOKPs}

\begin{table*}[htbp]
\centering
\caption{Averaged HV values obtained by CTAEA, RVEA, NSGA2, AGEMOEA, SMSEMOA, and the best operator generated by the proposed LLMOPT on 10 MOKPs over 10 independent runs. Superior averaged results (Lower HV values) on each instance are highlighted in bold font.}
\setlength{\tabcolsep}{8pt} 
\renewcommand{\arraystretch}{\tablestretch} 
\begin{tabular}{lcccccc}
\toprule
Problem & LLMOPT & CTAEA & RVEA & NSGA2 & AGEMOEA & SMSEMOA \\
\midrule
MOKP1 & \textbf{9.191e-03} & 1.771e-02 & 2.103e-02 & 1.895e-02 & 1.599e-02 & 1.705e-02 \\ 
MOKP2 & \textbf{5.268e-03} & 1.054e-02 & 1.241e-02 & 9.356e-03 & 8.762e-03 & 9.007e-03 \\ 
MOKP3 & \textbf{4.346e-03} & 1.215e-02 & 1.504e-02 & 1.170e-02 & 8.730e-03 & 1.039e-02 \\ 
MOKP4 & \textbf{8.762e-03} & 1.807e-02 & 2.224e-02 & 1.668e-02 & 1.462e-02 & 1.553e-02 \\ 
MOKP5 & \textbf{9.029e-03} & 2.167e-02 & 2.873e-02 & 1.916e-02 & 1.832e-02 & 1.861e-02 \\ 
MOKP6 & \textbf{6.690e-03} & 1.450e-02 & 1.810e-02 & 1.444e-02 & 1.286e-02 & 1.334e-02 \\ 
MOKP7 & \textbf{1.244e-02} & 2.265e-02 & 3.193e-02 & 2.183e-02 & 1.927e-02 & 1.981e-02 \\ 
MOKP8 & \textbf{3.490e-03} & 9.116e-03 & 1.285e-02 & 8.393e-03 & 7.468e-03 & 7.569e-03 \\ 
MOKP9 & \textbf{8.272e-03} & 1.523e-02 & 1.770e-02 & 1.405e-02 & 1.331e-02 & 1.329e-02 \\ 
MOKP10 & \textbf{4.780e-03} & 1.296e-02 & 1.644e-02 & 1.165e-02 & 1.114e-02 & 1.078e-02 \\ 
Averaged HV & \textbf{7.227e-03} & 1.546e-02 & 1.965e-02 & 1.462e-02 & 1.305e-02 & 1.354e-02 \\ 
\bottomrule
\end{tabular}
\label{tab:MOKP}
\end{table*}

\begin{figure*}[htbp]
\centering
\subfigure[MOKP-3]{
\includegraphics[width=0.45\columnwidth]{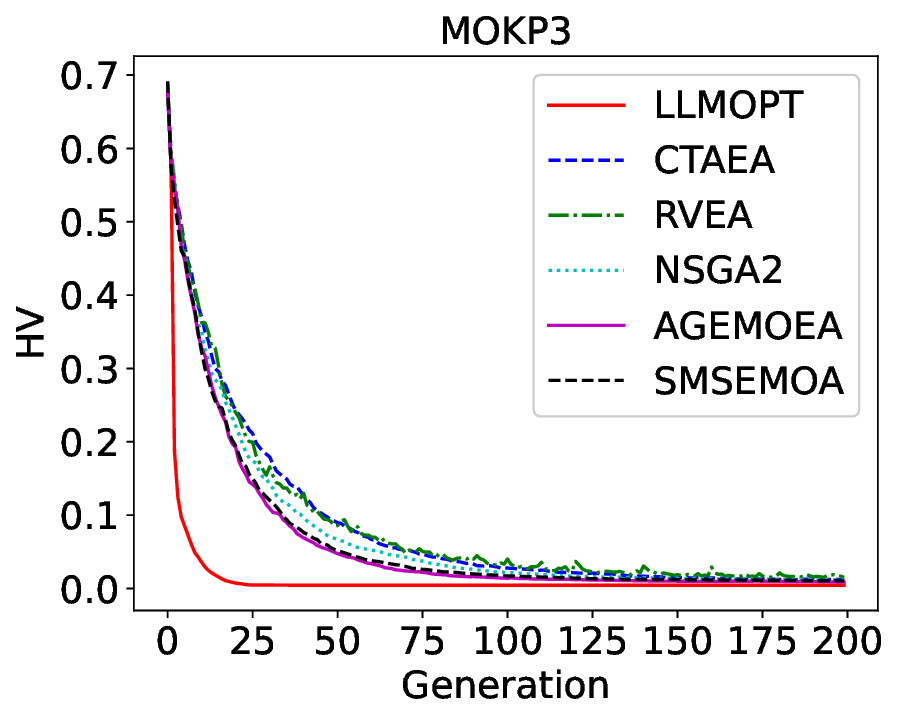}}
\subfigure[MOKP-5]{
\includegraphics[width=0.45\columnwidth]{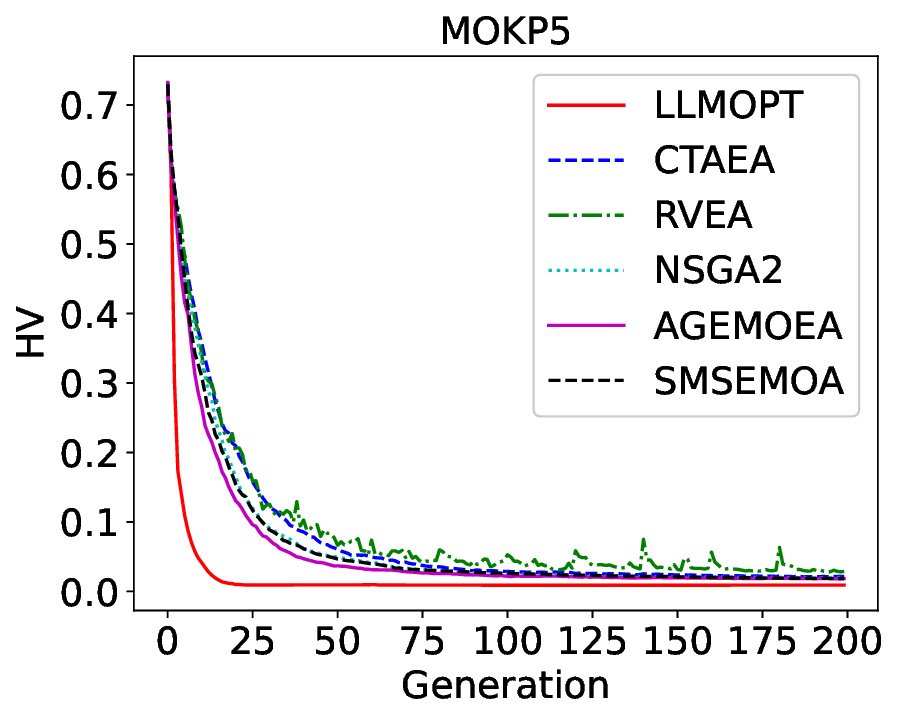}}
\subfigure[MOKP-6]{
\includegraphics[width=0.45\columnwidth]{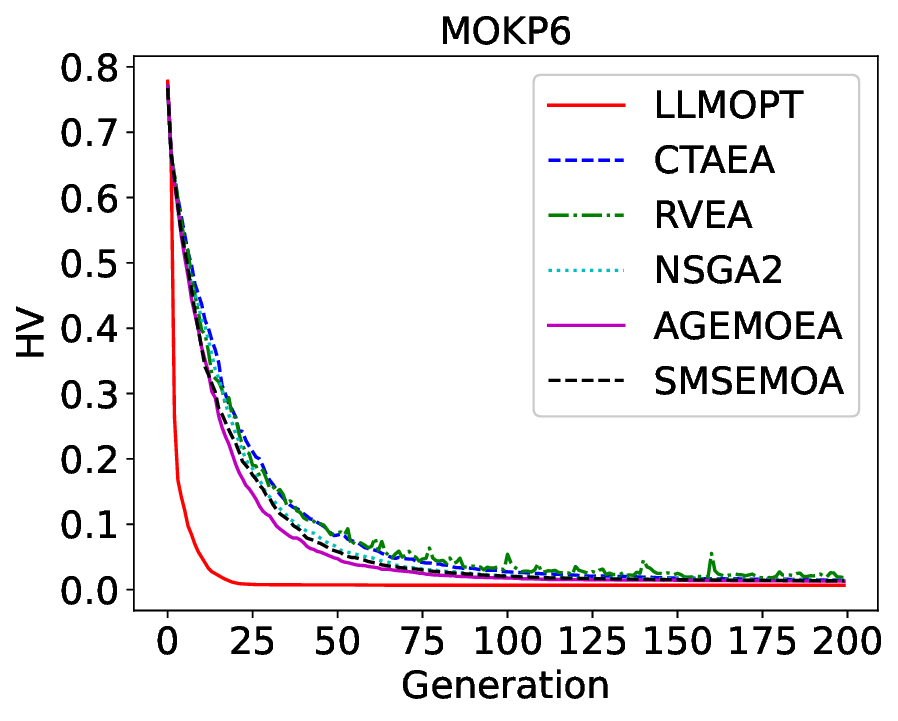}}
\subfigure[MOKP-8]{
\includegraphics[width=0.45\columnwidth]{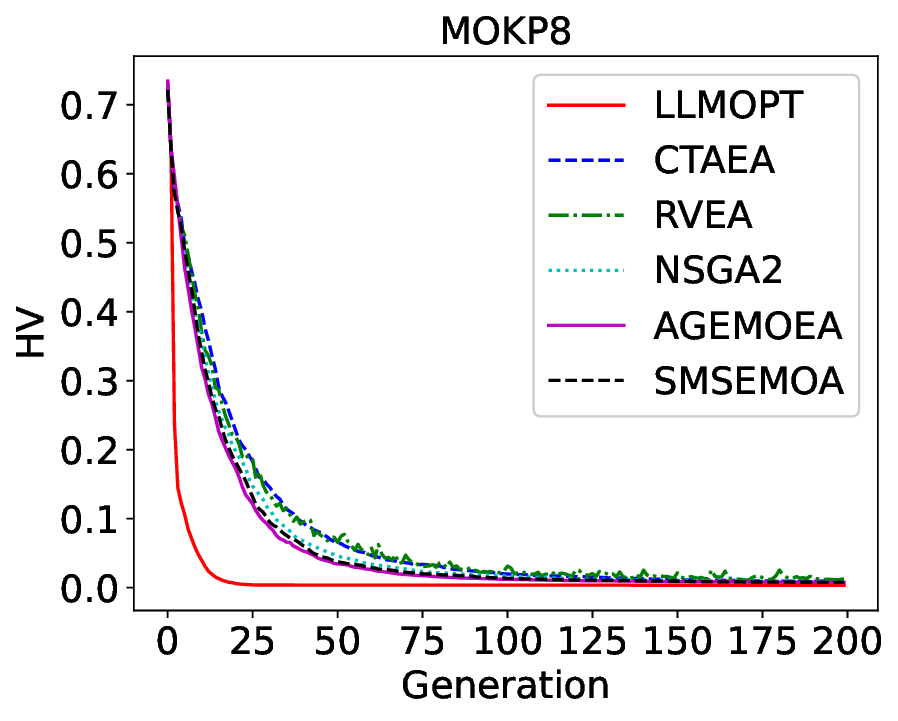}}
\caption{Convergence curves obtained by CTAEA, RVEA, NSGA2, AGEMOEA, SMSEMOA, and the best operator generated by the proposed LLMOPT on the representative MOKPs.}
\label{fig:MOKP}
\end{figure*}

In assessing the search performance on MOKPs, Table \ref{tab:MOKP} enumerates the average HV values achieved by CTAEA, RVEA, NSGA-II, AGEMOEA, SMSEMOA, and the optimal operator developed through the proposed LLMOPT. This evaluation encompasses 10 testing MOKP instances, each subjected to 10 independent runs. As can be observed, the `Average HV' row aggregates the HV metrics, providing a summarized view of the algorithms' overall performance across all testing instances. Examination of Table \ref{tab:MOKP} reveals that the EA operator gained by LLMOPT consistently outperforms the competing methods across all testing cases. This underscores the LLM's capacity to construct a potent evolutionary operator when provided with clearly defined problem attributes in the prompts. A closer look at the best operator generated by LLMOPT in the Supplementary Materials reveals an effective repair function crafted by the LLM, which strategically removes the heaviest item from the knapsack, thereby obtaining peak performance on the MOKPs.

Complementing the tabulated results, Fig.\ref{fig:MOKP} visually charts the convergence paths of the algorithms under comparison. The EA operator evolved by LLMOPT distinguishes itself by attaining the fastest convergence speed across all representative instances. The significant reduction in HV values not only validates the effectiveness of LLMOPT but also its robustness when face diverse multi-objective optimization challenges.

\subsubsection{Evaluation on MOTSPs}
Here we showcase the comparative performance analysis of the compared algorithms, including the best operator developed via our proposed LLMOPT, on the testing MOTSPs. Table \ref{tab:MOTSP} tabulates the average Hypervolume (HV) values achieved by MOEAD, CTAEA, RVEA, NSGA-II, AGEMOEA, SMSEMOA, and the operator developed via LLMOPT on testing MOTSP instances. The tabulated results demonstrate the efficacy of LLMOPT, particularly in its ability to generate effective EA operators for MOTSP. LLMOPT's operator outperforms the other algorithms in most instances, achieving the highest averaged HV values in MOTSP3 through MOTSP9. This is indicative of its superior capability to navigate the complex search space of MOTSPs effectively. While MOEAD exhibits slightly higher averaged HV values in MOTSP1 and MOTSP2, the operator's consistent performance across the majority of test cases suggests a more reliable and versatile approach to solving MOTSPs, which is also demonstrated by the bolded `Averaged HV' in the table.

\begin{figure*}[htbp]
\centering
\subfigure[MOTSP-1]{
\includegraphics[width=0.45\columnwidth]{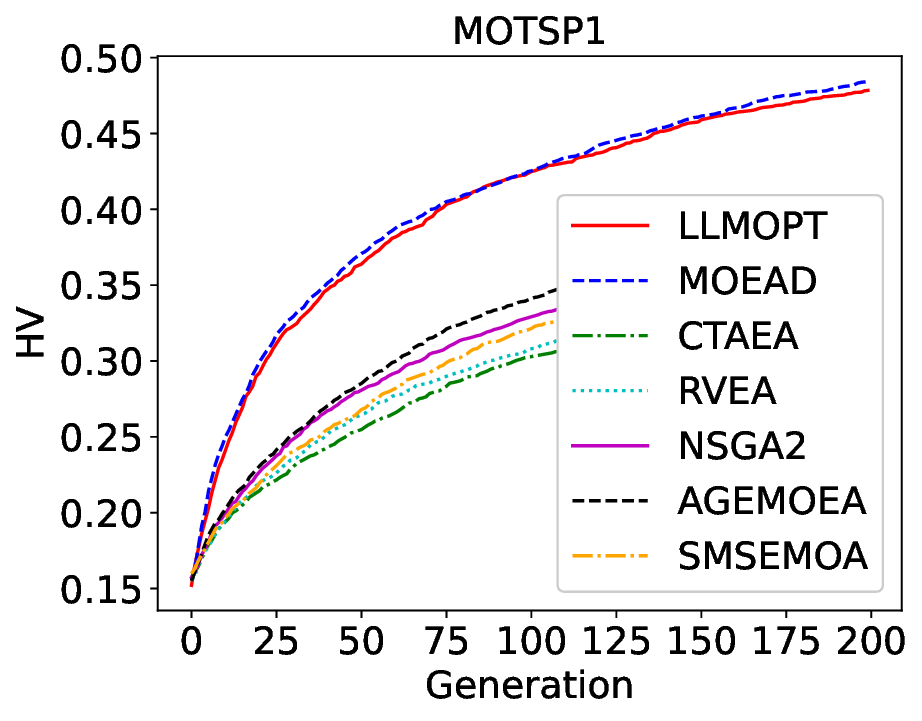}}
\subfigure[MOTSP-5]{
\includegraphics[width=0.45\columnwidth]{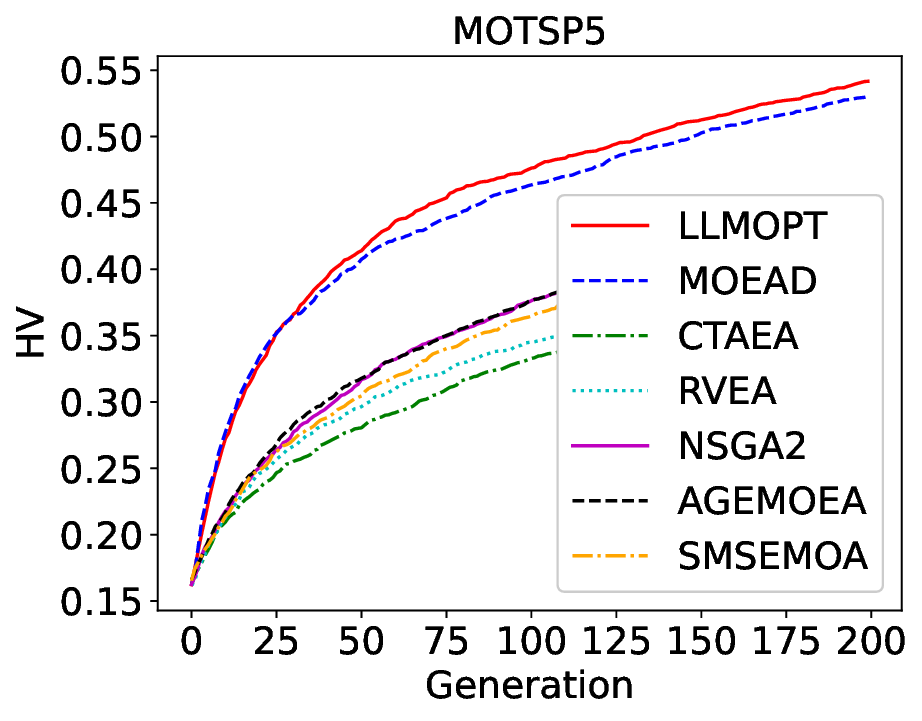}}
\subfigure[MOTSP-6]{
\includegraphics[width=0.45\columnwidth]{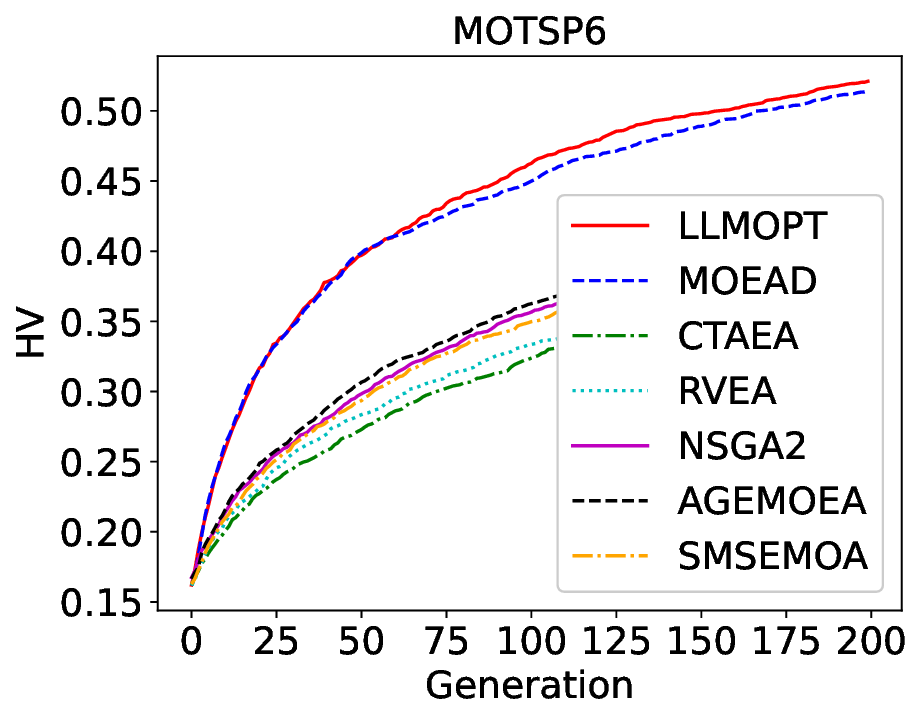}}
\subfigure[MOTSP-9]{
\includegraphics[width=0.45\columnwidth]{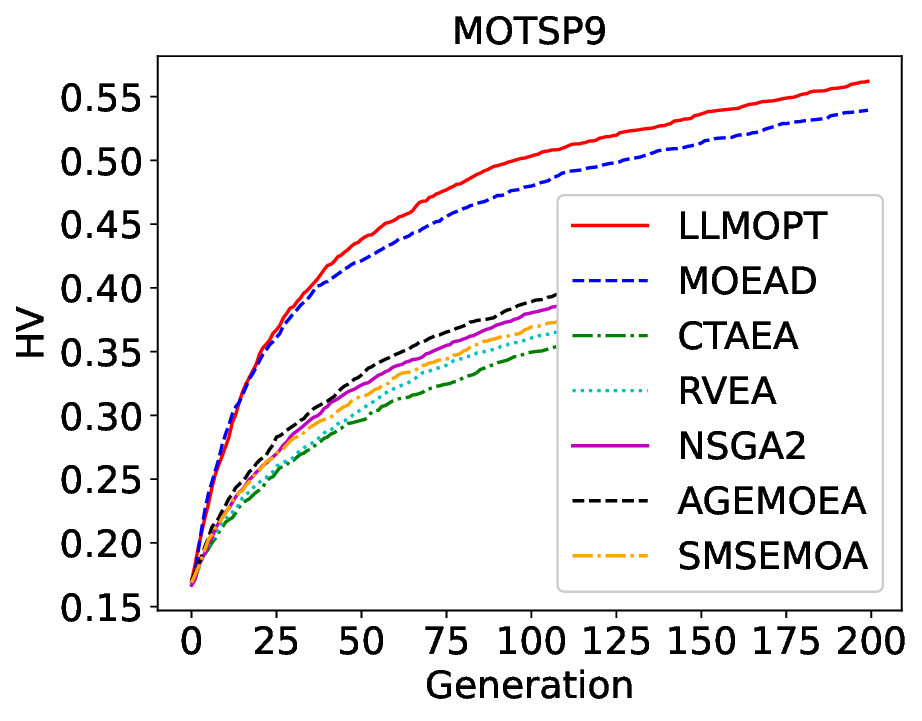}}
\caption{Convergence curves obtained by MOEAD, CTAEA, RVEA, NSGA2, AGEMOEA, SMSEMOA, and the best operator gained by LLMOPT on the representative MOTSPs.}
\label{fig:MOTSP}
\end{figure*}

Moreover, Fig. \ref{fig:MOTSP} illustrates the convergence trajectories for the compared algorithms on representative MOTSP instances. The X-axis represents the evolutionary progress through generations, while the Y-axis shows the average HV values from 10 independent runs. The visual data reveals that LLMOPT's best operator consistently achieves faster convergence, further demonstrating its capability to effectively solve a range of MOTSPs. This performance is attributed to the proposed LLMOPT, which automatically refine EA operators by leveraging problem-specific knowledge at hand. Delving into the Supplementary Materials for a more detailed examination of the optimal operator devised by LLMOPT, the LLMOPT utilizes a proficient crossover mechanism, specifically the edge recombination crossover (ERX) \cite{ahmed2010genetic}, to adeptly address the intricacies of MOTSPs.

\begin{table*}[htbp]
\centering
\caption{Averaged HV values obtained by MOEAD, CTAEA, RVEA, NSGA2, AGEMOEA, SMSEMOA, and the best operator generated by the proposed LLMOPT on 10 MOKPs over 10 independent runs. Superior averaged results (larger HV values) are highlighted in bold font.}
\setlength{\tabcolsep}{8pt} 
\renewcommand{\arraystretch}{\tablestretch} 
\begin{tabular}{lccccccc}
\toprule
Problem & LLMOPT & MOEAD & CTAEA & RVEA & NSGA2 & AGEMOEA & SMSEMOA \\
\midrule
MOTSP1 & 4.784e-01 & \textbf{4.844e-01} & 3.485e-01 & 3.583e-01 & 3.844e-01 & 3.954e-01 & 3.750e-01 \\ 
MOTSP2 & 5.030e-01 & \textbf{5.050e-01} & 3.669e-01 & 3.721e-01 & 3.983e-01 & 4.084e-01 & 3.991e-01 \\ 
MOTSP3 & \textbf{5.262e-01} & 5.198e-01 & 3.824e-01 & 3.961e-01 & 4.241e-01 & 4.322e-01 & 4.129e-01 \\ 
MOTSP4 & \textbf{6.082e-01} & 5.804e-01 & 4.462e-01 & 4.479e-01 & 4.980e-01 & 5.035e-01 & 4.847e-01 \\ 
MOTSP5 & \textbf{5.416e-01} & 5.303e-01 & 3.866e-01 & 4.012e-01 & 4.358e-01 & 4.420e-01 & 4.248e-01 \\ 
MOTSP6 & \textbf{5.211e-01} & 5.142e-01 & 3.736e-01 & 3.866e-01 & 4.132e-01 & 4.234e-01 & 4.090e-01 \\ 
MOTSP7 & \textbf{5.346e-01} & 5.175e-01 & 3.810e-01 & 3.846e-01 & 4.235e-01 & 4.358e-01 & 4.120e-01 \\ 
MOTSP8 & \textbf{5.256e-01} & 5.177e-01 & 3.753e-01 & 3.944e-01 & 4.174e-01 & 4.360e-01 & 4.113e-01 \\ 
MOTSP9 & \textbf{5.620e-01} & 5.393e-01 & 4.022e-01 & 4.207e-01 & 4.419e-01 & 4.502e-01 & 4.308e-01 \\ 
MOTSP10 & \textbf{5.160e-01} & 5.099e-01 & 3.722e-01 & 3.759e-01 & 4.115e-01 & 4.161e-01 & 3.956e-01 \\ 
Averaged HV & \textbf{5.317e-01} & 5.218e-01 & 3.835e-01 & 3.938e-01 & 4.248e-01 & 4.343e-01 & 4.155e-01 \\ 
\bottomrule
\end{tabular}
\label{tab:MOTSP}
\end{table*}

\subsubsection{Efficiency of Algorithms}

To evaluate the search efficiency of the EA operator generated via our proposed method, we focus on the execution time (wall clock time) as a primary metric, which is measured in seconds for a fair comparison across the algorithms MOEAD, CTAEA, RVEA, NSGA2, AGEMOEA, SMSEMOA. The investigation was conducted using testing groups of CMOP, MOKP, and MOTSP. 

\begin{table}[htbp]
\centering
\caption{Running time (wall clock time measured in seconds) obtained by MOEAD, CTAEA, RVEA, NSGA2, AGEMOEA, SMSEMOA, and the best operator optimized through LLMOPT on CMOP, MOKP and MOTSP. Smaller running time on each category is highlighted in bold font.}
\setlength{\tabcolsep}{12pt} 
\renewcommand{\arraystretch}{\tablestretch} 
\begin{tabular}{lrrr}
\toprule
& CMOP & MOKP & MOTSP \\
\midrule
LLMOPT & 32& \textbf{21}& 373 \\
MOEAD & 202& ---& 135 \\
CTAEA & 114& 79& 94 \\
RVEA & 37& 54& 64 \\
NSGA2 & \textbf{23}& 45& \textbf{53} \\
AGEMOEA & 50& 52& 60 \\
SMSEMOA & 1491& 48& 56 \\
\bottomrule
\end{tabular}
\label{tab:time}
\end{table}

The empirical evaluation of the search efficiency, as presented in Table \ref{tab:time}, indicates a significant variance in execution time across the competing algorithms. While the best operator generated via LLMOPT excels in MOKP, achieving the lowest running time of \textbf{21 seconds}, it falls behind in MOTSP, with a longer running time of 373 seconds, indicating the worst efficiency among the compared multi-objective algorithms. Despite this, the operator's efficacy in terms of solution quality for MOTSPs is superior, suggesting that its strategic approach, while time-intensive, results in high-quality solutions. In the domain of CMOPs, the operator's execution time is commendably efficient at 32 seconds, narrowly trailing behind NSGA2's \textbf{23 seconds}. This demonstrates the robustness of the proposed LLMOPT when handling different MOP categories.

\subsection{Investigation on Different LLMs}
\begin{figure}[htbp]
\centering
\includegraphics[width=0.8\columnwidth]{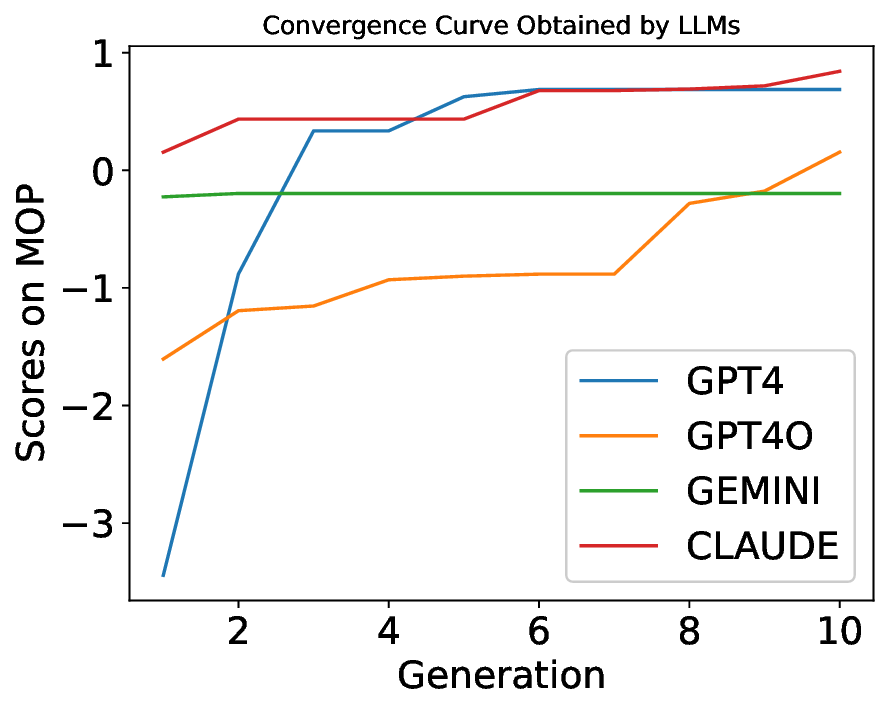}
\caption{Convergence curves gained by the proposed LLMOPT using different LLMs.}
\label{fig:investigate_llm}
\end{figure}

In this section, we rigorously validate the performance of our proposed LLMOPT framework using various LLMs. For brevity, we denote them as GPT4, GPT4O, GEMINI, and CLAUDE, respectively. Moreover, the same temperature (0.5) is adopted for each LLM. Fig. \ref{fig:investigate_llm} has illustrated the convergence curves obtained on CMOP validation dataset using these LLMs. As can be observed, different LLMs exhibit distinct behaviors within the proposed LLMOPT paradigm. Particularly, GPT4 and GPT4O initially struggle to produce a competent EA operator at the start of the search. However, through iterative optimization using LLMOPT, the effectiveness of operators generated by GPT4 and GPT4O improves significantly, highlighting the advantages of LLMOPT over traditional one-off program generation approaches. In contrast, CLAUDE consistently achieves the best EA operator performance from the outset until the end of the search. While GEMINI does yield a qualified EA operator initially, it fails to further enhance operator effectiveness over subsequent iterations. Based on these observations, the CLAUDE may have a superior performance in generating innovative EA operators against other LLMs.

\subsection{Ablation Study}
In this section, the ablation studies are conducted to demonstrate the efficacy of the proposed Pilot Run \& Repair component and dynamic selection strategy for LLM-based crossover.

\subsubsection{In-depth Analysis of Code Quality Among LLMs}

As we discussed in the introduction, a common challenge of using LLMs in program design is the occurrence of execution errors in the generated code, which can halt the iterative refinement process. To validate the efficacy of the proposed Pilot Run \& Repair component, we tested the code quality generated by GPT4, GPT4O, GEMINI and CLAUDE, using the initialization prompts. Particularly, each LLM was tasked to generate 20 EA operators, which were then assessed for their operational viability. During the testing phase, any errors encountered triggered a repair protocol, which is designed to systematically address and resolve coding errors. The results have been summarized in Table \ref{tab:error}, which presents a comparative analysis of the initial operator generation and subsequent repair success rates. As can be observed from, the `Tests' column records the number of attempts made to generate functional EA operators. The ``A + B'' within the `Success' column represents the total number of successfully generated EA operators, where ``A'' quantifies the instances without further interaction while ``B'' counts the successfully repaired instances via the proposed method. Moreover, `Failed' tallies the number of unrepaired operators after one trial, while `Repaired' reflects the proportion of successfully corrected operators through the repair process. The data presented in Table \ref{tab:error} clearly demonstrates the superior initial success rates of GPT4O and GEMINI, with an impressive 80\% of their generated operators being immediately deployable. Conversely, GPT4 shows a lower initial success but an outstanding repair success rate of 81.8\%, underscoring the `Pilot Run \& Repair' component's capability to effectively rectify errors. CLAUDE presents a balanced profile, showcasing a consistent performance in both initial operator generation and subsequent repairability.

\begin{table}[tbp]
\centering
\caption{Quality of operator generated with different LLMs}
\setlength{\tabcolsep}{12pt} 
\renewcommand{\arraystretch}{\tablestretch} 
\begin{tabular}{lrrrr}
\toprule
LLM & Tests & Success & Failed & Repaired (\%)\\
\midrule
GPT4 & 20 & 9 + 9 & 2 & 81.8\% \\
GPT4O & 20 & 16 + 2 & 2 & 50.0\% \\
GEMINI & 20 & 16 + 3 & 1 & 75.0\% \\
CLAUDE & 20 & 10 + 7 & 3 & 70.0\% \\
\bottomrule
\end{tabular}
\label{tab:error}
\end{table}

\begin{figure}[b]
\centering
\includegraphics[width=0.8\columnwidth]{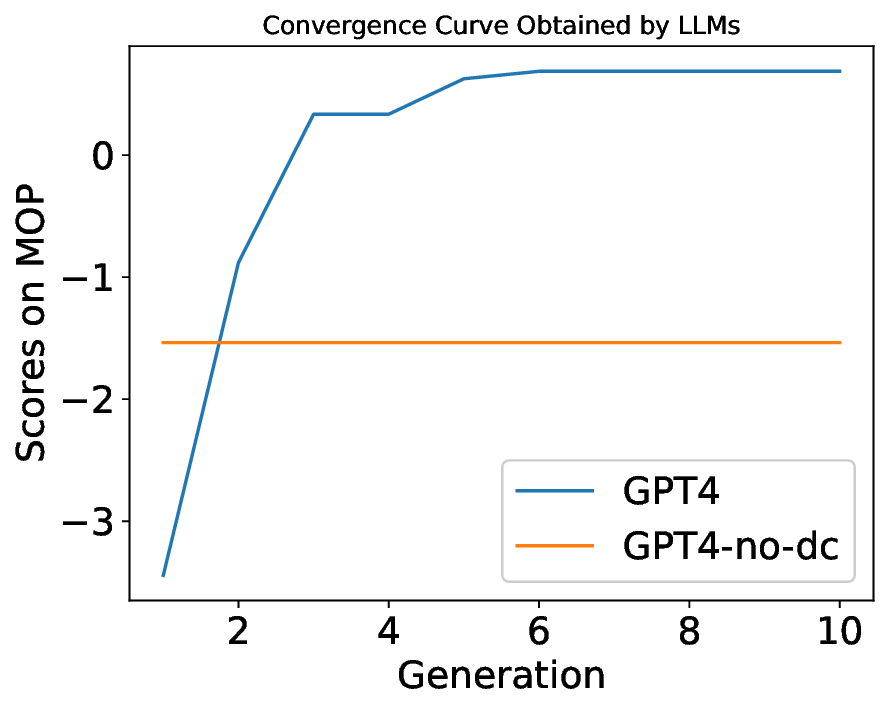}
\caption{Convergence curves gained by the proposed LLMOPT (denoted by `GPT4') and its counterpart without dynamic selection strategy (denoted by GPT4-no-dc).}
\label{fig:investigate_nodc}
\end{figure}

\subsubsection{Investigation on Dynamic Selection Strategy}
The selectiopn strategy plays an important role in evolutionary progresses. To validate the performance of the developed dynamic selection strategy, tailored for LLM-based textual optimization tasks, we present the convergence curves of the proposed LLMOPT (denoted by `GPT4') and its counterpart lacking this strategy (denoted by `GPT4-no-dc'). As can be observed in Fig. \ref{fig:investigate_nodc}, the `GPT4' model, equipped with the dynamic selection strategy, shows a significant improvement in terms of scores gained on CMOPs, indicating an effective adaptation and optimization capability brought by the developed dynamic selection strategy. In contrast, despite a good initialization for the operators, the `GPT4-no-dc' model's performance remains stagnant, again underscoring the necessity and impact of a dynamic selection strategy in the proposed LLMOPT.

\section{Conclusion} \label{sec:conclusion}

In this work, we have embarked on a journey to harness the capabilities of LLMs for the evolution of evolutionary operators, specifically targeting the nuanced domain of multi-objective optimization problems. Our proposed framework, LLMOPT, represents a paradigm shift in EA operator design, moving towards an autonomous methodology that minimizes the need for expert intervention. The 'Pilot Run \& Repair' mechanism embedded within LLMOPT has been instrumental in refining the evolution process of EA operators, enhancing the robustness of the generated operator and setting the stage for fully autonomous programming. The empirical studies showcases the superiority of LLM-evolved operators, demonstrating their advantage over traditional human-crafted approaches.

The future of this field is promising, with the potential for LLMs to further revolutionize the optimization landscape. Our results point to an emerging era of self-evolving, self-adapting, and self-improving programs, autonomously driven by the challenges they are engineered to conquer. The application of our framework to a broader spectrum of MOPs, particularly those of greater complexity and scale, stands as a significant next step.

\bibliography{IEEEabrv,traditional,LLM,MOPs}
\bibliographystyle{IEEEtran}

\end{document}